\documentclass[10pt,twocolumn,letterpaper]{article}

\usepackage[final]{cvpr}           % To produce the CAMERA-READY version
\usepackage{graphicx}
\usepackage{amsmath}
\usepackage{amssymb}
\usepackage{booktabs}
\usepackage{multirow}
\usepackage[accsupp]{axessibility} 
\usepackage[table,xcdraw]{xcolor}
\usepackage[utf8x]{inputenc} 

\usepackage[pagebackref,breaklinks,colorlinks]{hyperref}
\usepackage[numbers]{natbib}
\usepackage{notoccite}
\usepackage{xcolor}

\usepackage[capitalize]{cleveref}
\crefname{section}{Sec.}{Secs.}
\Crefname{section}{Section}{Sections}
\Crefname{table}{Table}{Tables}
\crefname{table}{Tab.}{Tabs.}

\def\confName{CVPR}
\def\confYear{2022}

\newcommand{\myparagraphn}[1]{%
\noindent\textbf{#1}\enspace}%
\newcommand{\myparagraph}[1]{%
\smallskip\myparagraphn{#1}}%
\usepackage{atbegshi,picture}
\usepackage{lipsum}

\author{%
Heming Yao$\footnotemark[1]$
\quad Phil Hanslovsky
\quad Jan-Christian Huetter \\
\quad Burkhard Hoeckendorf
\quad David Richmond$\footnotemark[1]$ \\
Biology Research $\mid$ AI Development (BRAID), gCS, Genentech \\ 
{\tt\small $*$ \{yao.heming, richmond.david\}@gene.com}
}

\AtBeginShipoutNext{\AtBeginShipoutUpperLeft{%
  \put(\dimexpr\paperwidth-2.3cm\relax,-1.5cm){\makebox[0pt][r]{\textcolor{gray}{This paper has been accepted for publication at the
IEEE Conference on Computer Vision and Pattern Recognition (CVPR)}}}%
}}

\AtBeginShipoutNext{\AtBeginShipoutUpperLeft{%
  \put(\dimexpr\paperwidth-5cm\relax,-2cm){\makebox[0pt][r]{\textcolor{gray}{
Workshop on Computer Vision for Microscopy Images (CVMI), Seattle, 2024. ©IEEE}}}%
}}

\begin{document}

\title{Weakly Supervised Set-Consistency Learning Improves Morphological Profiling of Single-Cell Images}

% Weakly Supervised Set-level Consistency Learning Improves Morphological Profiling from Optical Pooled Screening Data
\maketitle

%%%%%%%%% ABSTRACT
\begin{abstract}

Optical Pooled Screening (OPS) is a powerful tool combining high-content microscopy with genetic engineering to investigate gene function in disease. 
% by measuring single-cell morphology after the disruption of thousands of individual genetic elements within a single experimental well. 
% When combined with unbiased morphological markers, such as CellPaint, it can be used to profile thousands of genes for their role in a biological process of interest.
The characterization of high-content images remains an active area of research and is currently undergoing rapid innovation through the application of self-supervised learning and vision transformers.
% has inspired the development of multiple computer vision algorithms over the years. 
In this study, we propose a set-level consistency learning algorithm, Set-DINO, that combines self-supervised learning with weak supervision to improve learned representations of perturbation effects in single-cell images. Our method leverages the replicate structure of OPS experiments (\textit{i.e.}, cells undergoing the same genetic perturbation, both within and across batches) as a form of weak supervision. 
We conduct extensive experiments on a large-scale OPS dataset with more than 5000 genetic perturbations, and demonstrate that Set-DINO helps mitigate the impact of confounders and encodes more biologically meaningful information.
% amplifies perturbation-related morphological changes. In particular, compared to commonly used approaches, our proposed algorithm learns morphological features that have a higher replicate consistency across batches and encode more biologically meaningful information. 
In particular, Set-DINO recalls known biological relationships with higher accuracy compared to commonly used methods for morphological profiling, suggesting that it can generate more reliable insights from drug target discovery campaigns leveraging OPS.
% \textcolor{red}{impact statement}

\end{abstract}

%%%%%%%%% BODY TEXT
\section{Introduction}
\label{sec:intro}

High-content imaging combined with quantitative image analysis can be used to characterize cellular responses to genetic and chemical perturbations, and provides a powerful platform for target and drug discovery \cite{usaj2016high,cox2020tales,caicedo2016applications}.
Despite the prevalence of this approach in the pharmaceutical industry, arrayed screening still suffers from limitations due to the high cost of scaling to large genetic and chemical libraries.
Recently, Optical Pooled Screening (OPS), has been proposed as a cost-effective method for conducting high-content genetic screens at the whole-genome level \cite{feldman2019optical,walton2022pooled,sivanandan2023pooled,ramezani2023periscope}.
% on the other hand, is an emerging method that greatly reduces cost and increases scalability of high-content genetic screens, making it possible to conduct whole genome screens on multiple cell lines of interest 
% used for functional genomics studies and drug discovery . 
% This technique enables the introduction of thousands of genetic perturbations to a pool of cells in a single experiment. 
% Subsequently, in situ sequencing and fluorescence imaging techniques are used to identify the perturbations introduced to individual cells and capture single-cell morphological features at a low cost \cite{feldman2019optical}. 
% Its scalability and affordability make it possible to perform a large-scale exploration for cell lines of interest \cite{walton2022pooled}.
However, one caveat of pooled screens is that cellular phenotypes are captured at the single-cell level, in contrast to arrayed screens, where fields of hundreds of cells receive the same treatment.
This necessitates new research into methods for capturing cellular representations under perturbation that are robust to the high degree of noise and variability present in single-cell data.
 \cite{funk2022phenotypic,carlson2023genome,sivanandan2023pooled}.
% Previous works have demonstrated that quantitative measurements of cell morphology derived from OPS data can provide valuable insights in identifying genes across diverse cellular activities and inferring gene-gene relationships.

% The process starts with nuclei and cell segmentation, followed by the extraction of  classical image features such as intensity statistics, shape, size, and texture features.

% Many computer vision algorithms have been proposed to extract quantitative cellular representations from images.
CellProfiler remains one of the most widely used tools for extracting expert-defined features (also referred to as ``engineered" features) from high-content images \cite{carpenter2006cellprofiler, stirling2021cellprofiler}. 
% While CellProfiler shows good performance in characterizing cellular morphology \cite{funk2022phenotypic, way2022morphology}, the hand-crafted features are designed based on prior knowledge and may not be able to capture novel morphological features and features specific to a certain cell line. 
However, recent studies have focused on training deep learning algorithms, such as weakly supervised learning \cite{moshkov2024learning,caicedo2018weakly}, generative modeling \cite{carlson2023genome} and self-supervised learning (SSL) \cite{sivanandan2023pooled,doron2023unbiased,kim2023self,kraus2023masked}, to extract learned representations from high-content images.
Among those methods, DINO \cite{caron2021emerging} has emerged as a promising technique for extracting information-rich representations, and outperformed other approaches in a recent head-to-head comparison \cite{doron2023unbiased,kim2023self}.

While SSL frameworks are powerful feature extractors, they are unfortunately susceptible to learning unwanted confounding factors \cite{doron2023unbiased,kim2023self}. Such factors can include plate-to-plate variation, well-position effect, and experimental conditions, all of which can influence image intensity, contrast, and texture. Despite efforts toward optimal experimental design, confounding factors remain a persistent challenge in high-content screening. 
% SSLs have inherent limitations in simultaneous learning of both confounding factors and biologically meaningful features because their training objective emphasizes learning unbiased and rich representation from images.
Weakly supervised DINO \cite{cross2022self,haslum2024metadata} was proposed to address the sensitivity of SSL to confounders by sampling image pairs across experimental batches and thus encouraging DINO to learn batch-invariant representations. 
This approach has been shown to improve the quality of learned representations on downstream biological tasks \cite{cross2022self,haslum2024metadata,haslum2024bridging}; 
however, it has not yet been applied in the setting of single-cell images from optical pooled screens.
% this technique was proposed for images containing multiple cells derived from arrayed cell painting assays and hasn't been applied to single cell images yet, which present additional challenge due to a large degree of cell-to-cell variation.

In this study, we develop an SSL framework explicitly for single-cell images. 
Inspired by \cite{cross2022self,haslum2024metadata}, we adopt a cross-batch sampling strategy in an attempt to learn representations that are invariant to confounders.
However, we observed that DINO training with cross-batch sampling collapses due to the strong cell-to-cell variation exhibited in single-cell data.
% which can overwhelm the subtle perturbation effect.  This variation renders the cross-batch sampling strategy inapplicable to the DINO framework because enforcing consistent representations on single cells from different batches makes the model collapse to extract trivial representation, despite those cells undergoing the same perturbation.
To address this challenge, we combine cross-batch sampling with set-level representation to characterize cell populations undergoing a specific perturbation.

Our main contributions in this study include:

\begin{enumerate}
  \item We propose weakly supervised set-consistency learning (Set-DINO), a novel representation learning framework designed specifically for single-cell images from optical pooled screens.
  % This framework suppresses confounding factors and amplifies perturbation-related morphological features by leveraging the existence of numerous cell replicates undergoing the same perturbation, both within and across batches in OPS.
  To the best of our knowledge, this is the first time that set-level representation has been combined with DINO to facilitate self-supervised representation learning on noisy samples.

  \item We apply Set-DINO to a large-scale OPS dataset of more than 5000 essential genes where it achieves a significantly better performance compared to both engineered features and the standard DINO framework. Through extensive experiments, we demonstrate that Set-DINO leverages the weak supervision provided by cell replicates to extract cell representations that are both less sensitive to confounding factors and contain more biologically meaningful information. 
  % Thus, we anticipate that the proposed framework could benefit future target discovery research. 
  
\end{enumerate}

% Batch correction techniques have been used in many studies to mitigate unwanted variations in extracted morpholgocal features, and have achieved satisfying performance in integrating data from different batches \cite{doron2023unbiased,kim2023self,haslum2024bridging}. However, the dominance of technical variations in latent embedding during the model training can harm the training of self-supervised learning models. 

% TODO: cross-batch consistency learning was proposed for cell painting, it didn't work on single cell image
% we proposed set-level algorithm, emphasize (1) specifally for single cell images (2) first time combining set-theory with DINO.

\section{Related Work}

\subsection{Deep learning for high-content images}

% CellProfiler \cite{carpenter2006cellprofiler, stirling2021cellprofiler} has been successful in building morphological profiles on single cell images \cite{funk2022phenotypic, way2022morphology}. Despite displaying solid performance in characterizing cellular morphology, CellProfiler has several limitations. Hand-crafted features are proposed based on prior knowledge, and may not capture novel morphological features or cell line-specific features. Additionally, it relies upon accurate cell segmentation, which makes it less applicable when cells are challenging to segment.
%It begins with nuclei and cell segmentation, followed by the extraction of thousands of hand-crafted features from categories like intensity, shape, size, and texture. 

Deep learning has been extensively applied to the task of morphological profiling of high-content images.
For example, weakly supervised learning using perturbation labels was applied to images from arrayed screening in \cite{moshkov2024learning}, yielding improved performance at identifying treatments with the same mechanism of action (MoA) or genetic pathway, as compared to engineered features. 
However, leveraging perturbation labels as weak supervision may potentially result in learning spurious representations that falsely discriminate between different genetic perturbations with similar morphological effects. Not to mention the many genetic perturbations that have negligible effects on cell morphology.
% lead to reliance on spurious features because multiple genetic perturbations may result in similar morphological effects and many genetic perturbations have negligible effect on cell morphology. 

%Latent embeddings from auto-encoder have been analyzed in a genome-wide OPS \cite{carlson2023genome}, identifying several regulators of IRF3 translocation with higher accuracy than human-identified scores, despite translocation being a well-defined phenotype.

Self-supervised consistency learning provides an elegant solution to this problem, because it doesn't assume ``negative" relationships between samples from different perturbations.
Furthermore, SSL and Vision Transformers (ViT) have recently achieved state-of-the-art performance in learning representations from natural images that can generalize to downstream tasks \cite{caron2021emerging,he2022masked}.
% These methods have also been applied to high-content images. 
Sivanandan \textit{et al.} \cite{sivanandan2023pooled} and Doron \textit{et al.} \cite{doron2023unbiased} applied similar methods to high-content images, and demonstrated that embeddings from DINO with ViT led to improved performance. 
Masked auto-encoders (MAE) have also been shown to outperform weakly supervised baselines in uncovering biological relationships \cite{kraus2023masked}.
Further, in \cite{kim2023self}, several SSL techniques including SimCLR, DINO, and MAE were compared and DINO embeddings achieved the best performance in terms of reproducibility and target prediction for compound perturbations.

% Sivanandan \textit{et al.} and demonstrated that embeddings from DINO with ViT architecture on single cells led to improved gene interaction prediction compared with hand-crafted features and embeddings from self-supervised models pretrained on ImageNet \cite{sivanandan2023pooled}. Doron \textit{et al.} found that DINO with ViT encoded unbiased features of cellular morphology at multiple scales and could successfully infer biological relationships in various cellular imaging datasets \cite{doron2023unbiased}.

%In \cite{funk2022phenotypic}, CellProfiler features identified genes that regulate DNA repairs, cytoskeletal functions, and cell division. In \cite{way2022morphology}, CellProfiler features from cell painting data provided complementary information to gene expression profiling and captured biologically meaningful information for target prediction and the prediction of the mechanism of action for chemical compounds.

\subsection{Removing nuisance in morphological profiles}

Batch effect is a central challenge in learning biological meaningful representations from high-content imaging data.
Numerous batch correction techniques have been developed and successfully applied to mitigate unwanted variations in morphological profiles \cite{doron2023unbiased,kim2023self,haslum2024bridging}. 
However, those methods may not fully benefit deep learning approaches, as they are typically applied as a post-processing step.
% , whose learned representations tend to focus on strong technical variations, rather than the more subtle biologically meaningful signal.
% the dominance of technical variations in latent embedding during the model training can hinder the models to learn biologically meaningful information. 
Sypetkowski \textit{et al.} applied adaptive batch normalization to normalize features during training, using statistics from individual batches \cite{sypetkowski2023rxrx1}. Their proposed method mitigated batch effect and helped the model generalize to unseen batches. Inspired by this work, we explore an image normalization method based on image statistics from control cells. 

Furthermore, numerous methods leverage the replicate structure of high-content screening data to learn invariance to batch effects.
For example, Cross-Zamirski \textit{et al.} proposed a weakly supervised DINO model (WS-DINO) to incorporate the treatment labels in arrayed screening, and demonstrated improved performance in MoA prediction \cite{cross2022self}. 
Similarly, cross-domain consistency learning was proposed by Haslum \textit{et al.} with additional loss terms to force the model to disregard batch-specific signals \cite{haslum2024metadata}. Those methods follow a rationale similar to supervised contrastive learning \cite{khosla2020supervised}, where weak labels improve the robustness and informativeness of learned representations. 
Our study further validates the effectiveness of training SSL algorithms with weak labels, and extends this finding to single-cell images, requiring the use of a set-level loss. Our results indicate that while both weakly supervised learning and SSL approaches have their respective limitations, their combination can improve representations for high-content imaging data.

\subsection{Modeling population heterogeneity}

When analyzing high-content imaging data, profiles of biological replicates are typically aggregated to represent the average or median response of a cell population to each perturbation \cite{caicedo2017data}. 
Existing research also suggests that higher-order statistics such as the dispersion and covariance of features may provide additional information and improve performance on downstream tasks \cite{rohban2019capturing}.
By leveraging a set-level implementation, we benefit from a smoother, population-level loss, while retaining single-cell level profiles.

Deep Sets is a popular technique that offers a general framework for extracting representations from sets of objects \cite{zaheer2017deep}. The concept of set-level representation has been successfully integrated with SimCLR to improve unsupervised meta-learning performance on natural images \cite{lee2023self}. Dijk \textit{et al.} applied Deep Sets to pre-extracted single-cell profiles from CellProfiler, learning an aggregation function that down-weights noisy cells \cite{dijk2022learning}. The aggregation function was trained using weakly supervised contrastive learning, and the resulting profiles prioritize biological signals over batch effects.
In this study, instead of using hand-crafted features, we explore the possibility of combining set-level representation with end-to-end training of weakly supervised DINO to facilitate representation learning directly from raw images.

\section{Methods}

\begin{figure*}[!t]
    \centering
    \includegraphics[width=16cm]{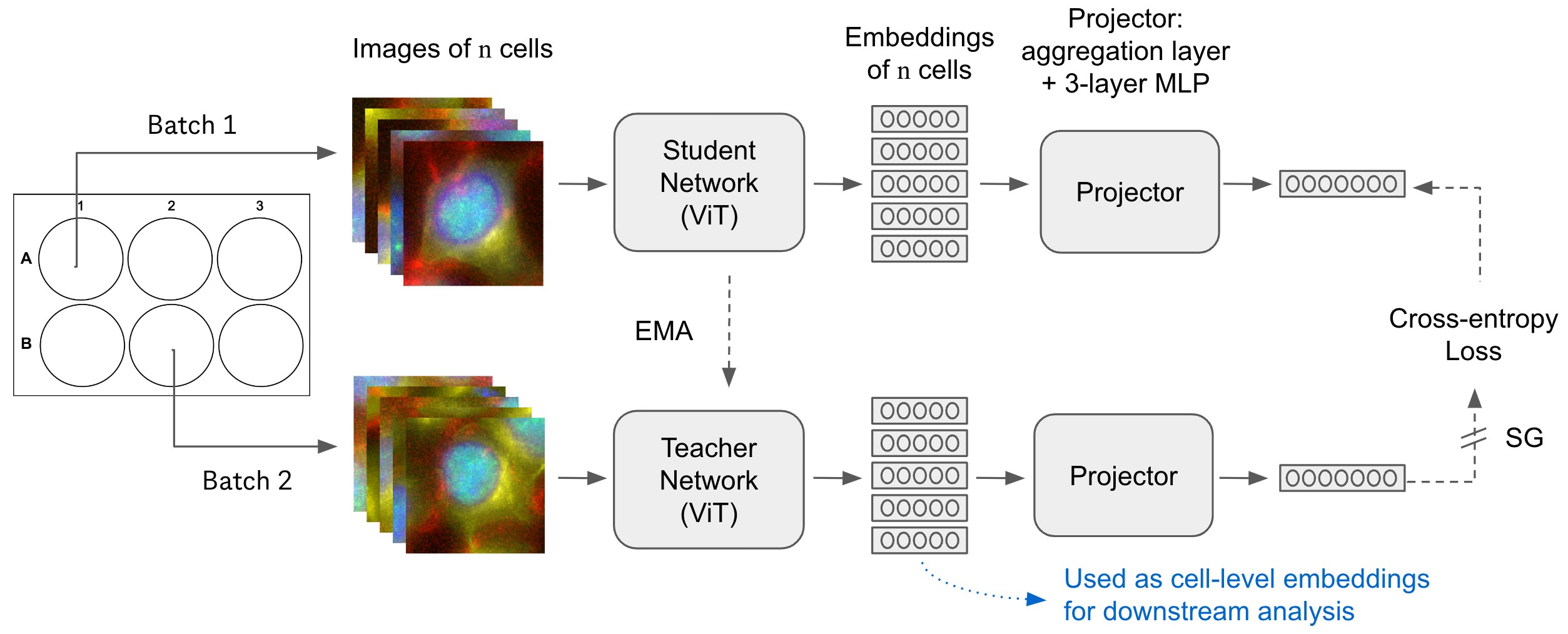}
    \caption{
\textbf{Overview of the Set-DINO framework.} The inputs are two sets of single-cell images undergoing the same perturbation in different batches. Each 4-channel image is processed individually by the Vision Transformer (ViT) to generate a set of embeddings. The projector consists of an aggregation layer, followed by three fully-connected layers. The resulting consensus embeddings from the student and teacher branches are used to calculate the cross-entropy loss to train the model. After the model is trained, the single-cell image embeddings from ViT are used as the cell-level morphological features. SG: stop-gradient, EMA: exponential moving average.
    }
    \label{fig:overview}
\end{figure*}

\subsection{Dataset}

We use a publicly released large-scale OPS dataset profiling CRISPR knockout of 5072 essential genes on cultured human cells \cite{funk2022phenotypic}. Four guide RNA (sgRNA) sequences were used per gene target, and an additional 250 non-targeting sgRNAs were used as negative controls. The entire sgRNA library was delivered to a pool of cells, and the experiment was replicated across 46 wells from 8 plates (each plate has at most 6 wells). 
% In-situ sequencing-by-synthesis was performed to identify the perturbation introduced to each cell, and phenotype 

In total, the dataset contains around 32 million cells, with a median of 6,000 cells per gene perturbation. 
The dataset was released with raw 4-channel images (stained for DNA, DNA damage, F-actin, and tubulin), metadata including the sgRNA that each cell received, and precomputed morphological features.
The released features are normalized by the median and median absolute deviation of non-targeting controls (NTCs) within the same well.

For the model training and evaluation, we divide the data from 8 plates into a training set (6 plates with 28 wells), a validation set (1 plate with 6 wells), and a test set (2 plates with 12 wells). The model is trained on the training set and the checkpoints and hyper-parameters are selected on the validation set. The test set is exclusively used to evaluate the model's performance and generalizability on unseen data. In this study, we regard each well as one experimental \textit{batch}. The median number of cells with the same sgRNA in each batch is 29.

\subsection{Image preprocessing}

We follow the established practice for preprocessing of high-content images \cite{janssens2021fully,ando2017improving,kim2023self}.
Images are flat-field corrected, and intensity values are clipped at the 0.1 and 99.9 percentiles, and then linearly re-scaled to $[0, 1]$. 
Single-cell images are then cropped using the cell centroids provided with the released metadata, and applying a 96-pixel-by-96-pixel bounding box around each cell. 

We evaluate two methods for normalizing the single-cell images. The first method is image-wise z-score normalization (referred to as \textit{z-score}), a common method where pixel intensities are image-wise and channel-wise normalized by z-score for each single-cell image \cite{janssens2021fully}. The second method is image normalization using statistics of the NTCs from the corresponding batch (referred to as \textit{NTC z-score}). In this approach, each single-cell image is normalized by the channel-wise mean and standard deviation of pixel intensities from all NTCs in the corresponding batch. 
We compare the performance of both normalization methods in our results section.
% According to our results, the second method helps mitigate batch effects.

% janssens2021fully: image-wise and channel-wise normalization
\subsection{Set-DINO framework}

Similar to the standard DINO \cite{caron2021emerging},  Set-DINO consists of a student branch and a teacher branch (\Cref{fig:overview}). In the standard DINO framework, a single image with different augmentations is fed into the student and teacher branches and the model is trained to maximize the similarity between the embeddings from the two branches. 

In Set-DINO, we sample a set of $n$ single-cell images from cells receiving perturbation $p$ in batch $b$: $\mathrm{X}_{p,b} = \{x^{1}_{p,b}, \ldots, x^{n}_{p,b}\}$. The $i$th tensor $x^{i}_{p,b} \in \mathbb{R}^{C, H, W}$ represents a multi-channel image of one cell receiving perturbation $p$ from batch $b$, where $C, H, W$ denote the number of channels, height, and width of the image, respectively. Similarly, we sample a second set of $n$ single-cell images from cells receiving the same perturbation $p$ from a different batch $b'$: $\mathrm{X}_{p,b'} = \{x^{1}_{p,b'}, \ldots, x^{n}_{p,b'}\}$.

The image sets $\mathrm{X}_{p,b}$ and $\mathrm{X}_{p,b'}$ are fed into the student network $\phi_{s}$ and teacher network $\phi_{t}$, respectively, to generate single-cell latent embeddings. Then, an aggregation layer $\Lambda$ aggregates the image-level embeddings to set-level embeddings: 
\begin{align}
\pi_{p,b} &= \Lambda({\phi_s(x^{1}_{p,b}), ..., \phi_s(x^{n}_{p,b})}),\\
\pi_{p,b'} &= \Lambda({\phi_t(x^{1}_{p,b'}), ..., \phi_t(x^{n}_{p,b'})}).
\end{align}

The embeddings $\pi_{p,b}$ and $\pi_{p,b'}$ represent a consensus of the cell populations receiving perturbation $p$ in batches $b$, and $b'$, respectively. They are used as a pair of views whose similarity is optimized during training of the student network $\phi_{s}$:
\begin{equation}
\mathrm{\mathcal{L}} = \mathrm{H}(\gamma(\pi_{p,b}), \gamma(\pi_{p,b'})),
\end{equation}
where $\gamma$ is a multi-layer perceptron (MLP), and $\mathrm{H}$ is the cross-entropy loss.

Similar to the standard DINO framework, a stop-gradient (SG) operator is applied on the teacher network $\phi_{t}$ and the parameters in the teacher network are updated with an exponential moving average (EMA) of the student parameters.
We use a ViT backbone for $\phi_{s}$ and $\phi_{t}$, and a 3-layer MLP for $\gamma$. 
The aggregation function $\Lambda$ can be any function that is invariant to permutations such as feature statistics \cite{lee2023self}, Deep Sets \cite{zaheer2017deep}, or self-attention layers \cite{lee2018set}. In this study, we use the arithmetic mean because of its simplicity and effectiveness \cite{lee2023self}.

The motivation of the set-level representation is to better characterize the cell population within a specific condition, while retaining cell-to-cell variation. 
Moreover, we found that the set-level representation is necessary for stabilizing DINO training when applying a cross-batch sampling strategy, due to the large degree of variation in single-cell images. 

The cross-batch sampling strategy can be regarded as a form of data augmentation using biological replicates. Compared to common image augmentation techniques such as rotations and Gaussian blur, the utilization of cell replicates from different batches serves as a more biologically meaningful form of augmentation. However, despite having received the same genetic perturbation, single cells sampled from different batches may demonstrate very different morphology due to variations in cell states, cell cycle, and batch-level technical variations. This is especially true considering that the perturbation effects from many genetic perturbations can be extremely subtle \cite{moshkov2024learning}. Moreover, due to the varying effectiveness of the guide RNA, some cells may ``escape" the perturbation, and exhibit a morphology similar to NTCs. Consequently, contrasting two single-cell images sampled from different batches may make the network insensitive to both batch effects and biologically meaningful information. Our experiments in this study demonstrate that without set-level aggregation, the model collapses, as excessively strong data augmentation forces the model to extract very general features that are identical for all cells in the dataset. To address this problem, we create views from each experimental batch using a set of cells receiving the same perturbation. The two sets of cells are assumed to contain similar distributions in cell states.

In this study, we explore different cell sampling strategies.
% For cells receiving the same genetic perturbation, different sgRNAs may result in a similar directional shift in morphology but with different magnitudes due to variations in on-target efficiency and off-target activity \cite{labitigan2022mapping}. 
Given that every gene target in our dataset has four sgRNAs, we compare sampling cells receiving the same sgRNA versus those receiving perturbation of the same gene target. As an ablation study, we also experiment with sampling sets of cells from the same experimental batch.

\begin{table*}[h]
\small
\centering
\caption{\textbf{Evaluation of batch-level gene profiles and consensus gene profiles.} We compare the performance of profiles from the Set-DINO framework with the standard DINO framework and engineered features from \cite{funk2022phenotypic}. ``Set-DINO - sgRNA" means that the teacher and student views are sampled from cells with the same guide RNA, while ``Set-DINO - gene target" means that the views are sampled from cells with the same gene target (multiple guide RNAs). ``KNN@k=5" refers to the accuracy of the k-nearest neighbor classifier when $k=5$. All values are displayed in percentages. Best values are highlighted in bold.}
\label{table:comparison}
\begin{tabular}{@{}c|cc|cc|cc|cc@{}}
\toprule
\multicolumn{1}{c|}{}                   & \multicolumn{2}{c|}{}                                                                                                           & \multicolumn{2}{c|}{}                                                                    & \multicolumn{4}{c}{{\color[HTML]{000000} Biological Recall}}                                                                                                                                                                                                  \\ \cline{6-9} 
\multicolumn{1}{c|}{\multirow{-2}{*}{}} & \multicolumn{2}{c|}{\multirow{-2}{*}{Batch Effect}}                                                                             & \multicolumn{2}{c|}{\multirow{-2}{*}{Reproducibility}}                                   & \multicolumn{2}{c|}{CORUM}                                                                                                              & \multicolumn{2}{c}{CORUM (curated)}                                                                                 \\ \hline
\multicolumn{1}{c|}{}                   & \begin{tabular}[c]{@{}c@{}}↓ KNN\\ @k=5\end{tabular} & \multicolumn{1}{c|}{\begin{tabular}[c]{@{}c@{}}↓ GC\\ @k=5\end{tabular}} & \begin{tabular}[c]{@{}c@{}}↑ KNN\\ @k=5\end{tabular} & \multicolumn{1}{c|}{↑ mAP}        & \begin{tabular}[c]{@{}c@{}}↑ Recall\\ @5\%\end{tabular} & \multicolumn{1}{c|}{\begin{tabular}[c]{@{}c@{}}↑ Recall\\ @10\%\end{tabular}} & \begin{tabular}[c]{@{}c@{}}↑ Recall \\ @5\%\end{tabular} & \begin{tabular}[c]{@{}c@{}}↑ Recall\\ @10\%\end{tabular} \\ \midrule

Engineered features                               & 19.2                                                 & 31.8                                                & 2.58                                                 & 1.62          & 25.9                                                    & 33.0                                                     & 27.9                                                     & 35.6                                                    \\ \midrule
DINO (z-score)                             & \textbf{12.6}                                                 & \textbf{10.1}                                                & 0.68                                                 & 0.41          & 21.1                                                    & 28.4                                                     & 23.5                                                     & 31.3                                                     \\
DINO (NTC z-score)                         & 21.7                                                 & 34.5                                                & 1.16                                                 & 0.64          & 25.4                                                    & 33.7                                                     & 28.4                                                     & 36.9                                                     \\ \midrule
Set-DINO (z-score) - sgRNA           & 22.5                                                 & 42.8                                                & 5.48                                                 & 3.55          & 27.3                                                    & 46.2                                                     & 33.8                                                     & 43.9                                                     \\
Set-DINO (NTC z-score) - sgRNA       & 20.2                                                 & 34.4                                                & 5.71                                                 & 3.71          & 28.6                                                    & 37.9                                            & 35.0                                                     & 45.3                                                     \\
Set-DINO (z-score) - gene target & 19.2                                        & 32.1                                       & 6.18                                       & 4.10 & 28.7                                           & 37.3                                                     & 35.2                                            & 45.5                                            \\
Set-DINO (NTC z-score) - gene target & 17.3                                        & 23.4                                       & \textbf{6.87}                                        & \textbf{4.51} & \textbf{29.5}                                           & \textbf{38.3}                                                     & \textbf{36.1}                                            & \textbf{46.9}                                            \\ \bottomrule
\end{tabular}
\end{table*}

% \begin{figure*}[h]
%     \centering
%     \includegraphics[width=16cm]{figure-feature-agg.png}
%     \caption{
% \textbf{Morphological feature processing workflow.} Raw features are single-cell profiles from the trained ViT or engineered features from \cite{funk2022phenotypic}. The median and median absolute deviation of features from non-targeting cells in each batch are calculated for batch-wise normalization and centering. Morphological profiles of cells with the same gene target from the same batch are aggregated into batch-level gene profiles. Subsequently, consensus gene profiles are generated by aggregating batch-level gene profiles across all replicate batches.
%     }
%     \label{fig:workflow}
% \end{figure*}

\subsection{Implementation}

In Set-DINO, the data loader samples cellular images based on perturbation labels. To build one mini-batch, we initially select $N_P$ perturbations, followed by sampling a pair of batches for each perturbation. Subsequently, for all cells in a given batch $b$ with perturbation $p$, we randomly sample $n$ cells to build the image set $X_{p,b}$. We experiment with $n \in \{1, 4, 8, 16\}$, and maintain $N_P \times n = 512$ to maximize GPU utilization. Each epoch consists of 50k mini-batches.

We use ViT-small/16 as the backbone and set the hidden dimension of the MLP to 2048. The model is trained with an Adam optimizer for 300 epochs. We follow the same warm-up and cosine schedule for learning rate and weight decay as in the standard DINO framework \cite{caron2021emerging}, with a base learning rate of 0.04. The teacher temperature is set to 0.01. Eight local crops are used for each single-cell image. The Set-DINO framework is implemented in PyTorch and distributed over 2 GPUs. With $n=16$, the model training takes 12 days.

We make the Set-DINO framework and the checkpoint of a trained model publicly available. \footnote{\href{https://github.com/Genentech/set-dino}{https://github.com/Genentech/set-dino}}

%\toprule

\subsection{Representation levels}

After model training, single-cell embeddings from ViT are extracted, processed and then aggregated into multiple levels for further evaluation \cite{caicedo2017data}.

\myparagraph{Single-cell profiles:} For Set-DINO and standard DINO models, the embeddings of the class token from the last four ViT layers are used as cell-level features. As a baseline, we also use the engineered features released by \cite{funk2022phenotypic}. The raw features are normalized using batch-wise normalization based on the median and median absolute deviation of features from NTCs, aiding in data alignment across different batches and mitigating batch effects \cite{caicedo2017data}. Experimentally, we find that engineered features benefit from Principal Component Analysis (PCA), but learned features do not. As a result, we apply PCA to engineered features (after normalization) with a cutoff of 95\% variance. 
% Thus, single-cell profiles represent normalized cell-level morphological features.

\myparagraph{Batch-level gene profiles:} Single-cell profiles of cells with the perturbations of the same gene target from the same batch are aggregated by an arithmetic mean operation into batch-level gene profiles. Batch-level gene profiles represent the cell population with a specific genetic perturbation from a specific batch.

\myparagraph{Consensus gene profiles:} Batch-level gene profiles are batch-wise centered on the means of features from NTCs and subsequently aggregated across batches to generate consensus gene profiles. Consensus gene profiles represent the average morphological changes resulting from individual genetic perturbations. These embeddings can be used to infer gene functions and gene-gene relationships.

\begin{figure*}[!t]
    \centering
    \includegraphics[width=17cm]{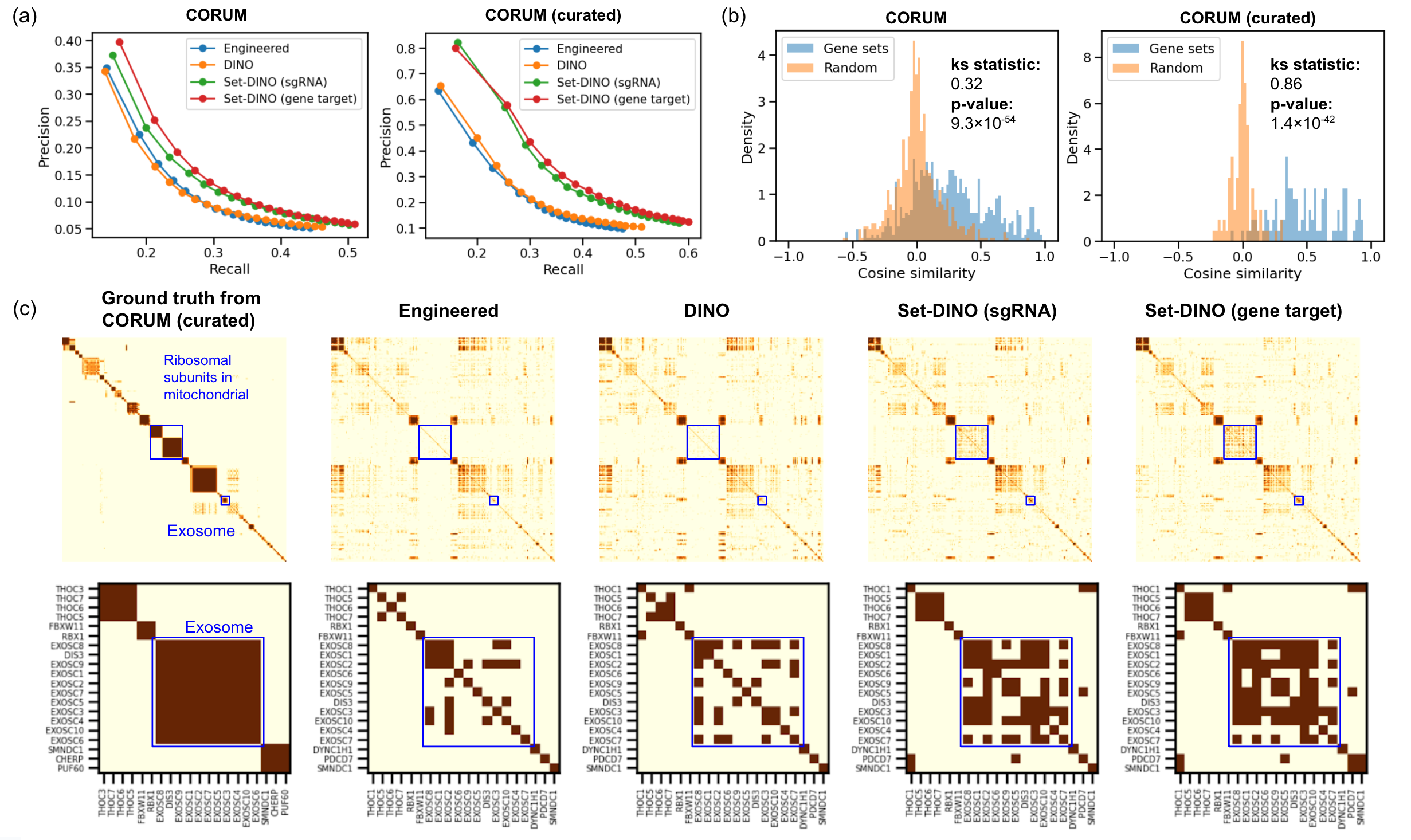}
    \caption{
\textbf{Evaluation of consensus gene profiles in identifying gene-gene relationships.} (a) Precision-recall curves for predicting gene-gene relationships according to CORUM and curated CORUM. The curves are drawn by varying cutoff percentiles from top 1\% to top 20\%. (b) Distribution of cosine similarities between gene pairs. ``Gene Sets" contains gene pairs that are involved in the same protein complex, while ``Random" contains randomly sampled gene pairs. (c) The first row contains the full adjacency matrices of the ground truth graph (leftmost column) and prediction graphs (other columns). The second row provides a zoomed-in view of the adjacency matrix focused around the exosome. Although self-edges are included in the visualization, they are not considered when calculating biological recall and precision. All DINO and Set-DINO models shown in this figure were trained with NTC z-score normalization. 
    }
    \label{fig:corum}
\end{figure*}

\subsection{Evaluation protocols}

We employ multiple metrics to evaluate our learned representations on the basis of reproducibility, batch effect and biological recall.

\myparagraph{Reproducibility:} After feature processing and aggregation, we first evaluate the reproducibility of the batch-level gene profiles, following previous evaluation approaches \cite{kim2023self,chandrasekaran2022three}.
Specifically, a graph is constructed where the nodes are batch-level gene profiles and the edge weights are given by the cosine distance between every pair of nodes. Then, we compute the average precision for the task of predicting the genetic perturbation of each node from the genetic perturbations of its distance-ranked neighbors, which measures the ability to retrieve the profiles of the same genetic perturbations from different batches against the background of all other perturbations. 

Following this, the mean average precision (mAP) is calculated across all nodes. Also, we calculate the k-nearest neighbors (KNN) classification accuracy on perturbations when $k=5$. A high mAP and accuracy indicate that batch-level gene profiles from the same genetic perturbation are clustered and dissimilar to other genetic perturbations as well as NTCs. Given that many perturbations exhibit negligible effects and those cells post-perturbation display very similar profiles to NTCs, the absolute values of mAP and accuracy are not expected to be high.

\myparagraph{Batch Effect:} To evaluate the batch effect, we calculate the KNN classification accuracy on experimental batches using the same graph described above ($k=5$). In addition, we compute the Graph Connectivity (GC) \cite{kim2023self} on the KNN graph. To calculate GC, subgraphs are constructed by retaining only nodes from a certain batch. GC is then defined as the average ratio of the number of nodes in the largest connected component and the total number of nodes in the subgraph. Low batch prediction accuracy and GC values indicate that the embeddings of cells from different experimental batches are well mixed (i.e., low batch effect).

\myparagraph{Biological Recall:} Finally, we evaluate the biological information in consensus gene profiles by measuring how well they can infer gene-gene relationships \cite{celik2022biological}. The biological ``ground truth" is built from the CORUM database \cite{giurgiu2019corum}, a public collection of manually curated mammalian protein complexes. A ground truth graph is built by connecting every pair of genes in the same protein complex. A prediction graph is constructed by connecting every pair of genes whose cosine similarity between the morphological profiles exceeds a certain percentile of the pairwise similarity distribution. The prediction graph is compared with the ground truth graph, and the recall of the gene-gene relationships is calculated using the top 5\% percentile and top 10\% percentile \cite{celik2022biological} as cutoffs. In addition, a precision-recall curve is calculated for further evaluation.

In CORUM, some protein complexes significantly overlap with others, potentially causing the involved genes to dominate the ground truth graph. To avoid this bias, we utilized a curated CORUM database from \cite{funk2022phenotypic}, which includes only protein complexes with limited overlap with other complexes. The full CORUM contains gene-gene relationships from 1263 genes that are perturbed in our OPS dataset, and the curated CORUM includes a subset of 538 genes.

% \begin{figure*}[!t]
%     \centering
%     \includegraphics[width=17cm]{figure-corum.png}
%     \caption{
% \textbf{Evaluation of consensus gene profiles in identifying gene-gene relationships.} (a) Precision-recall curves for predicting gene-gene relationships according to CORUM and curated CORUM. The curves are drawn by varying cutoff percentiles from 80\% to 99\%. (b) Distribution of cosine similarities between gene pairs. ``Gene Sets" contains gene pairs that are involved in the same protein complex, while ``Random" contains randomly sampled gene pairs. (c) The first row contains the full adjacency matrices of the ground truth graph (leftmost column) and prediction graphs (other columns). The second row provides a zoomed-in view of the adjacency matrix focused around the exosome. Although self-edges are included in the visualization, they are not considered when calculating biological recall and precision. All DINO and Set-DINO models shown in this figure were trained with NTC z-score normalization. 
%     }
%     \label{fig:corum}
% \end{figure*}

\begin{table*}[t]
\small
\centering
\caption{\textbf{Ablation study on cell sampling strategies.} We compare the performance of three sampling strategies with different numbers of cells, $n$. 
In the \textit{same cell(s)} strategy the teacher and student views are built from the same set of cells. When $n=1$, this is equivalent to the standard DINO framework.
In the \textit{within-batch} strategy 
the views are built from two sets of cells from the same batch with the same perturbation (same guide).
In the \textit{cross-batch} strategy the views are built from two sets of cells from different batches with the same perturbation (same guide).
% All sampling is done at the guide level.
All values in this table are displayed in percentages.
Values are omitted for models trained with the \textit{cross-batch} strategy and $n\in {1,4}$ because their training collapsed. All DINO and Set-DINO models shown in this table were trained with NTC z-score normalization on input images. ``KNN@k=5" refers to the accuracy of the k-nearest neighbor classifier when $k=5$.
Best values are highlighted in bold.
}
\label{table:sampling_strategy}
\begin{tabular}{@{}c|cc|cc|cc|cc@{}}
\toprule
\multicolumn{1}{c|}{}                   & \multicolumn{2}{c|}{}                                                                                                           & \multicolumn{2}{c|}{}                                                                    & \multicolumn{4}{c}{{\color[HTML]{000000} Biological Recall}}                                                                                                                                                                                                  \\ \cline{6-9} 
\multicolumn{1}{c|}{\multirow{-2}{*}{}} & \multicolumn{2}{c|}{\multirow{-2}{*}{Batch Effect}}                                                                             & \multicolumn{2}{c|}{\multirow{-2}{*}{Reproducibility}}                                   & \multicolumn{2}{c|}{CORUM}                                                                                                              & \multicolumn{2}{c}{CORUM (curated)}                                                                                 \\ \hline
\multicolumn{1}{c|}{}                   & \begin{tabular}[c]{@{}c@{}}↓ KNN\\ @k=5\end{tabular} & \multicolumn{1}{c|}{\begin{tabular}[c]{@{}c@{}}↓ GC\\ @k=5\end{tabular}} & \begin{tabular}[c]{@{}c@{}}↑ KNN\\ @k=5\end{tabular} & \multicolumn{1}{c|}{↑ mAP}        & \begin{tabular}[c]{@{}c@{}}↑ Recall\\ @5\%\end{tabular} & \multicolumn{1}{c|}{\begin{tabular}[c]{@{}c@{}}↑ Recall\\ @10\%\end{tabular}} & \begin{tabular}[c]{@{}c@{}}↑ Recall \\ @5\%\end{tabular} & \begin{tabular}[c]{@{}c@{}}↑ Recall\\ @10\%\end{tabular} \\ \midrule
Engineered features       & \textbf{19.2}                                        & \textbf{31.8}                                       & 2.58                                                 & 1.62          & 25.9                                                    & 33.0                                                     & 27.9                                                     & 35.6                                                         \\ \midrule
Same cell(s), n=1  & 21.7                                                 & 34.5                                                & 1.16                                                 & 0.64          & 25.4                                                    & 33.7                                                     & 28.4                                                     & 36.9                                                     \\
Same cell(s), n=4  & 24.2                                                 & 45.8                                                & 1.00                                                 & 0.54          & 22.9                                                    & 30.9                                                    & 27.3                                                     & 36.1                                                     \\ \midrule
Within-batch, n=1  & 47.1                                                 & 75.7                                                & 0.97                                                 & 0.53          & 21.9                                                    & 30.5                                                     & 26.3                                                     & 36.1                                                     \\
Within-batch, n=4  & 44.2                                                 & 71.8                                                & 0.47                                                 & 0.28          & 19.8                                                    & 29.0                                            & 24.9                                                     & 34.3                                                     \\ \midrule
Cross-batch, n=1,4 & ---                                          & ---                                         & ---                                         & ---   & ---                                            & ---                                                       & ---                                            & ---                                            \\
Cross-batch, n=8   & 20.0                                                 & 36.2                                                & \textbf{5.83}                                                 & \textbf{3.78}          & 27.7                                                    & 36.4                                                     & 34.7                                                     & 45.1                                                     \\
Cross-batch, n=16  & 20.2                                                 & 34.4                                                & 5.71                                        & 3.71 & \textbf{28.6}                                           & \textbf{37.9}                                            & \textbf{35.0}                                            & \textbf{45.3}                                            \\ \bottomrule
\end{tabular}
\end{table*}

\section{Results and Discussion}
\subsection{Set-DINO achieves superior performance compared to existing methods}

\Cref{table:comparison} shows our results on reproducibility and batch effect on batch-level gene profiles, as well as the performance of gene-gene relationship inference using consensus gene profiles. The reproducibility metrics and batch effect metrics should be considered together to assess the quality of the learned representation. The former evaluates the replicate consistency and how well the model captures the perturbation effect, while the latter evaluates the model's resistance to batch-level confounding factors.

These results show that the standard DINO model with NTC z-score yields a performance similar to engineered features in predicting gene-gene relationships. Notably, Set-DINO significantly outperforms both of these on the gene-relationship task using both CORUM and curated CORUM. With the optimal Set-DINO gene profiles, the recall at top 5\% cutoff increases by 8.2\% (29.4\% relative increase) on curated CORUM compared to engineered features. Additionally, the reproducibility metrics for Set-DINO profiles markedly surpass those of both DINO and engineered features. This boost suggests that Set-DINO's weakly supervised training encourages the model to learn more biologically meaningful representations.

Interestingly, within the self-supervised setting, the inference performance of gene-gene relationships correlates with our reproducibility metric. This suggests that combining weak supervision on perturbation labels with SSL leads to morphological features that encode biologically meaningful information more effectively. 

Furthermore, comparing the guide-level (sgRNA) and gene-level variants of Set-DINO, we observe that building views from all cells with the same gene target leads to a lower batch effect, and a relative increase of over 20\% on reproducibility metrics, and a slight increase in predicting gene-gene relationships.

Finally, we observe a consistent improvement in all performance metrics when using NTC z-score normalization. For standard DINO with z-scoring, we note that although its batch effect metrics are low, it also exhibits low replicate consistency, which indicates that this model has less capability to extract distinguishable features. Finally, our batch effect metrics only consider batch-level confounding factors. Therefore, it is possible that the embeddings from DINO with z-score are dominated by other nuisance factors, such as cell positions within the well. 

\subsection{Set-DINO encodes biologically meaningful information}
We further illustrate the biological information encoded in consensus gene profiles in \Cref{fig:corum}. \Cref{fig:corum}a compares recall-precision curves from different methods using cutoff ranges from top 1\% to top 20\%, focusing on the high-precision regime that is required for target discovery. We note that while standard DINO and engineered features exhibit similar performance, Set-DINO achieves a substantial performance boost, especially on curated CORUM.

We also observe that gene pairs with known relationships tend to have higher cosine similarity than randomly sampled gene pairs (\Cref{fig:corum}b). The KS-statistic between the two distributions is 0.32 on CORUM, and 0.86 on curated CORUM.

\Cref{fig:corum}c presents the adjacency matrices from the ground truth graph on curated CORUM, as well as the predicted graphs from different methods. We find that consensus gene profiles from Set-DINO models achieve higher recall in multiple protein complexes. Notably, the gene-gene relationships in some protein complexes are almost entirely missed in engineered and DINO profiles but captured by Set-DINO. Two examples of this are ``ribosomal subunits in mitochondria", which are critical for mitochondrial translation \cite{sastre2000mitochondria}, and exosomes, which play an integral role in cell-cell communication \cite{kalluri2020biology}. Both of these structures are highlighted in \Cref{fig:corum}c.

\subsection{Ablation studies on Set-DINO framework}

We perform a series of ablation studies to evaluate the effect of weak supervision in self-supervised learning of cellular embeddings. See \Cref{table:sampling_strategy} for a summary of these results.

Our first observation is that model training tends to collapse if we apply the proposed cross-batch sampling strategy with a small value of $n$. This phenomenon is likely due to the substantial differences in cell state distribution and the general batch-level distribution shift exhibited by the two sets of cells. Increasing the value of $n$ aids in stabilizing the model during training. Based on our results, increasing beyond $n=8$ has a negligible impact on performance.

In addition to cross-batch sampling, we propose two alternative sampling strategies. The first strategy is the ``same cell" approach, in which the identical sets of cells are used for both teacher and student branches. When $n = 1$, this is equivalent to the standard DINO framework. A comparison of the results from $n = 1$ and $n = 4$ reveals that utilizing a set of images leads to higher batch effects and lower reproducibility metrics. The recall in gene-gene relation prediction is also lower. These results are likely due to the averaging done in the aggregation layer, which may reduce the effect of image augmentation.

The second ``within-batch" strategy involves building teacher and student views by sampling different cells with the same perturbation from the same batch. In this case, the differences between the two views mainly arise from the variation in cell state distributions due to random sampling. The results indicate that the profiles from within-batch sampling exhibit markedly worse batch effects as well as lower reproducibility metrics and gene-gene relationship prediction performance. This suggests that in this scenario, the model extracts primarily batch-related information to ensure consistency between the teacher and student branches. 

\begin{figure}[!t]
    \centering
    \includegraphics[width=8.5cm]{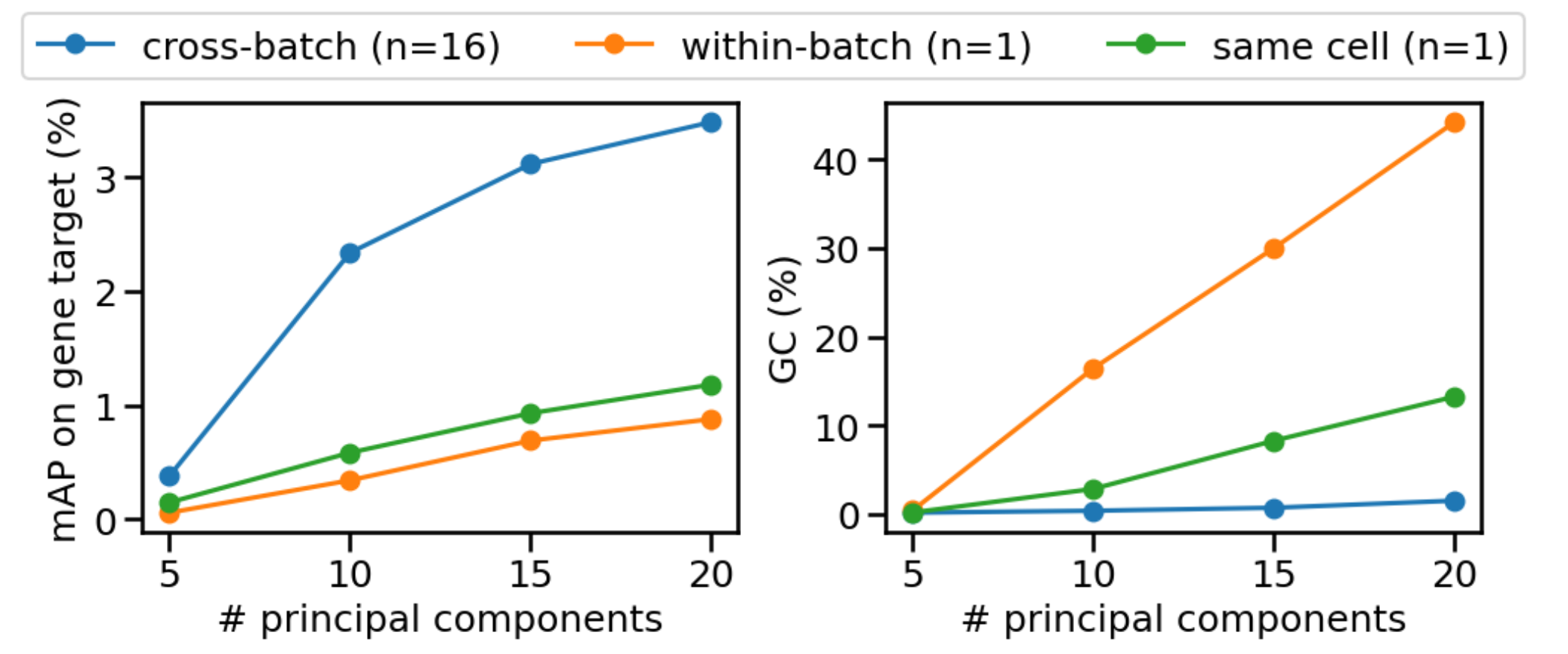}
    \caption{
\textbf{Performance analysis on an increasing number of principal components.} Principal component analysis (PCA) is performed on batch-level gene profiles. Reproducibility and batch effect metrics on an increasing number of principal components are measured.
    }
    \label{fig:ablation}
\end{figure}

For additional validation, we perform a PCA on the batch-level gene profiles, and calculate the reproducibility and batch effect metrics on an increasing number of Principal Components (PCs) by order. This analysis provides insight into the amount of perturbation-specific and batch-specific signals contained in the PCs with the highest variances.
\Cref{fig:ablation} shows that the batch-specific signals dominate the high-ranked PCs of gene profiles from the model trained using the ``within-batch" strategy, while perturbation-specific signals dominate the high-ranked PCs from the ``cross-batch" trained model. In addition, according to the metrics we observed on the validation set during model training for within-batch sampling (not shown), batch-specific signals begin to dominate at an early stage in the training process.

%Upon increasing the value of $n$, the performance of the "within-batch" strategy converges to that of the "same cell" strategy, mainly because the two views share more cells when $n$ is high.

\section{Conclusion}

In this study, we propose the Set-DINO model with a cross-batch sampling strategy that combines weak supervision and self-supervised learning to obtain better single-cell representations for cell morphology images. Our results demonstrate that the proposed framework outperforms established baselines using engineered features and the standard DINO framework in extracting morphological profiles of single-cell images from a held-out test set. We conduct ablation studies to confirm that both set-level representation and cross-batch sampling are critical to achieving success. 

Additional evaluation based on prior biological knowledge reveals that the consensus gene profiles learned by Set-DINO significantly improve the prediction of gene-gene relationships. Thus, we anticipate that the proposed framework could benefit future target discovery and drug discovery research. Furthermore, while this study focuses on single-cell imaging data from optical pooled screens, Set-DINO may also be applicable to other single-cell datasets containing weak labels, as well as single-cell crops from arrayed cell painting datasets.
% to extract single-cell level morphological profiles. 

One limitation of this study is that a simple arithmetic average is used in the aggregation layer to fuse the latent embeddings from single-cell images to population-level representations. Previous studies have explored more sophisticated aggregation methods for computing set-level representations \cite{zaheer2017deep,lee2023self,lee2018set}. Future work will investigate which strategy is optimal for feature aggregation. 

\section*{Acknowledgments}
We would like to thank Avtar Singh and Luke Funk for productive discussions, and their support with the OPS dataset.

%%%%%%%%% REFERENCES
{\small
\bibliographystyle{ieee_fullname}
\bibliography{bib}
}

\end{document}